# Exploring the Distribution Regularities of User Attention and Sentiment toward Product Aspects in Online Reviews

Chenglei Qin[1] , Chengzhi Zhang[1,*], Yi Bu[2]

1. Department of Information Management, Nanjing University of Science and Technology, Nanjing, China
2. Department of Information Management, Peking University, Beijing, China

## Abstract

**Purpose -** To better understand the online reviews and help potential consumers, businessmen, and product manufacturers effectively obtain users' evaluation on product aspects, this paper explores the distribution regularities of user attention and sentiment toward product aspects from the temporal perspective of online reviews.

**Design/methodology/approach -** Temporal characteristics of online reviews (purchase time, review time, and time intervals between purchase time and review time), similar attributes clustering, and attribute-level sentiment computing technologies are employed based on more than 340k smartphone reviews of three products from JD.COM (a famous online shopping platform in China) to explore the distribution regularities of user attention and sentiment toward product aspects in this article.

**Findings -** The empirical results show that a power-law distribution can fit user attention to product aspects, and the reviews posted in short time intervals contain more product aspects. Besides, the results show that the values of user sentiment of product aspects are significantly higher/lower in short time intervals which contribute to judging the advantages and weaknesses of a product.

**Research limitations** – The paper can't acquire online reviews for more products with temporal characteristics to verify the findings because of the restriction on reviews crawling by the shopping platforms.

**Originality/value** –This work reveals the distribution regularities of user attention and sentiment toward product aspects, which is of great significance in assisting decision-making, optimizing review presentation, and improving the shopping experience.

## 1. Introduction

The numbers of online reviews of the MacBook Air in JD.COM and Kindle in Amazon had been over 200K and 60K, respectively, in the mid of 2018. It is difficult for potential shoppers, businessmen and manufacturers to find high-quality reviews that indicate the attitudes of consumers on products from numerous online reviews. Indeed, information overload in online reviews has been a significant obstacle for people to get valuable information (Ghose and Ipeirotis, 2010; Moghaddam and Ester, 2012; Mohammad and Dan, 2016; Yin *et al.*, 2014). Fortunately, researchers have explored in-depth in the fields of fake/spam reviews detection (e.g., Cardoso *et al.*, 2018; Jindal and Liu, 2008; Lau *et al.*, 2012; Li *et al.*, 2011; Li *et al.*, 2014) and reviews helpfulness assessing (e.g., Ghose and Ipeirotis, 2010; Karimi and Wang, 2017; Malik and Hussain, 2018; Ngo-Ye *et al.*, 2018; Zhang and Lin, 2018) that can alleviate information redundancy in a certain extent.

---

* Corresponding Author.

However, users' experience needs are constantly increasing. Potential consumers, businessmen, and product manufacturers are more concerned about users' evaluation on product aspects (Akpoyomare *et al.*, 2012; Rachim and Setiawan, 2014; Xiao *et al.*, 2018). Alam *et al.* (2016) found that users frequently looked for important aspects of a product or service in the reviews.

Generally, online reviews contain lots of product aspects. A product aspect with a more significant sentiment tendency would have more reference value. How to find these valuable reviews? To help people effectively obtain useful evaluation information on product aspects, this paper utilizes the temporal characteristics of online reviews and review mining technologies to exploring the distribution regularities of user attention and sentiment toward product aspects. The temporal characteristics of online reviews mainly include purchase time, review time (posting time), time intervals between purchase time and review time. An example of temporal characteristics of a product review is shown in Figure 1.

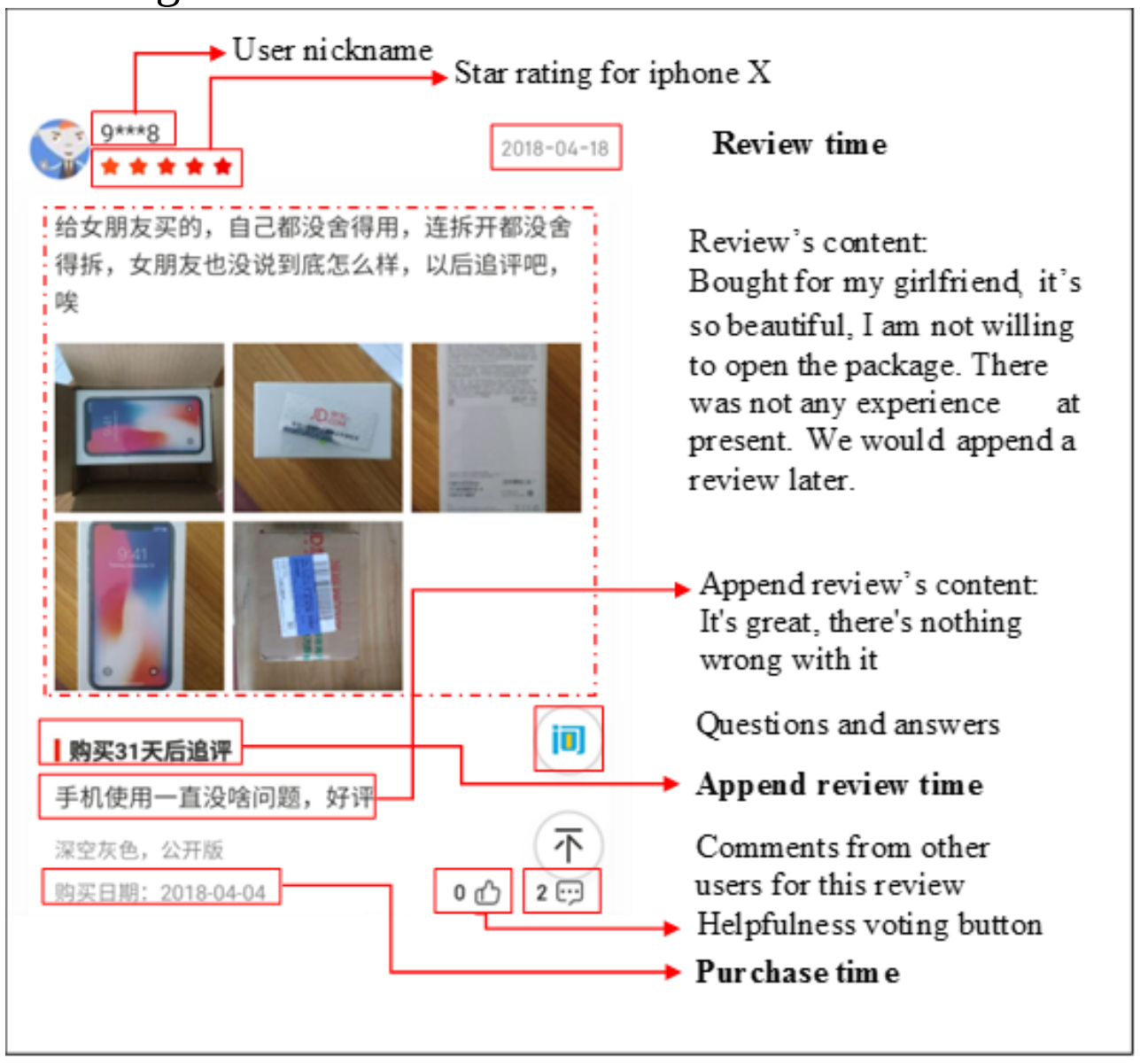

**Figure 1.** An example of temporal characteristics of a review of iPhone X from JD.COM

The paper explores the distribution regularities of user attention and sentiment toward product aspects based on more than 340k online reviews from three parts:

- Investigating the distribution regularity of user attention to product aspects at different time intervals;
- Exploring the distribution regularity of the number and types of product aspects at different purchase time and different time intervals;
- Analysing the distribution regularity of user sentiment on product aspects at different purchase time and different time intervals.

The major contributions of the work are as follows:

- As we know, this work is the first one investigating the distribution regularities of user attention and sentiment toward product aspects from the perspective of temporal characteristics of online reviews.
- The paper identifies the feasibility of utilizing temporal characteristics of online reviews, which reveals the distribution regularities of user attention and sentiment toward product aspects.
- The paper finds that the richness of online reviews' content is associated with users' review time. Online reviews posted in short time intervals will contain more aspects and have significant sentiment tendency, which will be conducive to fully understand the performance and judge the strength and disadvantages of a product.

- The findings of this work are of great significance for consumers, businessmen, and manufacturers in assisting decision-making, optimizing reviews presentation, and improving shopping experience.

The paper is structed as follows. In Section 2, the paper mainly introduces the research status of attribute-level sentiment analysis and the application of temporal characteristics in online reviews mining. Section 3 introduces the method used in this work. Section 4 analyses the experiment results. Section 5 discusses the significance of the distribution regularities. Section 6 concludes the paper.

# 2. Related Work

## *2.1 Attribute-Level Sentiment Computing*

Document, sentence, and attribute are three levels on which sentiment analyses are built (Tang *et al.*, 2009; Tang *et al.*, 2015; Yousif *et al.*, 2017; Zhang and Liu, 2011). The focus of this paper is the distribution pattern of user attention and user sentiment toward product attributes. Attribute extraction and attribute sentiment computing are the two subtasks of attribute-level sentiment analysis.

### *2.1.1 Attribute Extraction*

Product attribute extraction is an important branch in the field of reviews mining. The methods of attribute extraction of online reviews can be classified into three major classes, namely unsupervised, semi-supervised, and supervised (Rana and Cheah, 2016). In the early period, a representative unsupervised method was proposed by Hu and Liu (2004) who used the Apriori algorithm (Agrawal and Srikant, 1994) to automatically extract high-frequency nouns or noun phrases and removed the incorrect attributes with pruning rules. Popescu and Etzioni (2005) found the components of a product by calculating the score of mutual information between the candidate noun and the entity that has the whole-part relationship. Ku *et al.* (2006) used TF-IDF to calculate the importance of words in documents and sections to extract attributes. Blair-goldensohn *et al.* (2008) considered nouns or noun phrases that appear with sentiment words as candidate attributes. Ivan et al. (2008) proposed an unsupervised method that utilizing the LDA model and the pLSA model to extract product attributes form online reviews. Moghaddam and Ester (2010) proposed a method based on extra information of online reviews, such as predefined attributes and their ratings, as well as rating guideline, to extract attributes. However, this method may not work well in current shopping websites because of the huge differences of extra information among websites. Semi-supervised methods were also widely used. For example, Wu *et al.* (2009) and Yu *et al.* (2011) used Phrase Dependency Parsing technology and support vector machine (SVM) to extract product attributes. Yet, the performance of these methods depends on the results of parsing. Alam *et al.* (2016) proposed a domain-independent topic sentiment model called Joint Multi-grain Topic Sentiment (JMTS) to extract semantic aspects.

From the perspective of supervised methods, Jin *et al.* (2009) employed the hidden Markov model and Bootstrapping algorithm to extract camera attributes from online reviews; they argued that their proposed strategy could recognize new features without relying on natural language processing technologies. Zhang *et al.* (2009) and Xu *et al.* (2011) used conditional random fields, domain dictionary, and language features to extract product aspects, but it takes more time to construct and update the domain dictionary. Moreover, topic models (Andrzejews *et al.*, 2009; Xu *et al.*, 2015) and real-time review streams (Dragoni *et al.*, 2018) have also been used in attributes extraction.

There would be several similar descriptions for one product aspect. Numerous attributes of one product will bring difficulties to analyze users' evaluation on product aspects. Thus, similar product attributes need to be clustered. Several studies have been done on this topic. Li *et al.* (2006), for example, clustered similar attributes manually from movie commentary. Samaha *et al.* (2014) and Banken *et al.* (2014) used a dictionary to cluster similar attributes. Zhai *et al.* (2011) found that unsupervised learning methods based on distribution similarity need to be improved and, as a result, proposed to use a semi-supervised method to cluster similar attributes with a small amount of tagged corpus.

### *2.1.2 Attribute Sentiment Computing*

The sentiment within a sentence or a paragraph might not be consistent (Ma *et al.*, 2018). Thus, it is quite necessary to analyze people's attitudes on products in a fine-grained level. Yi *et al.* (2003) used sentiment lexicon, namely General Inquirer (GI), Dictionary of Affect of Language (DAL), and WordNet, to analyze the sentiment of product features. Ding *et al.* (2008), for example, utilized sentiment dictionary and language rules to quantify the sentiment polarity of product attributes. As a typical dictionary-based method, it is easy to execute. Similarly, Thet *et al.* (2010) used SentiWordNet and grammatical dependency to calculate clause sentiments corresponding to an attribute of movies. He *et al.* (2011) utilized generic sentiment lexicon and pseudo-labeled documents to train sentiment classifier. The performance of their model was better than weakly-supervised sentiment classification models. And, Ayoub *et al.* (2013) utilized polarity lexicon to analyze the attributes sentiment. Brody and Elhadad (2010) present an unsupervised system for determining the attribute sentiment in online reviews. The system first extracted the relevant adjectives, built a conjunction graph, automatically determined the seed set, and then propagated the polarity scores to the rest adjectives. Jo (2011) proposed Sentence-LDA (SLDA) model and extended SLDA to incorporate aspect and sentiment together to compute different aspects sentiment. Fu *et al.* (2013) used LDA to extract the global attributes from online social reviews and then utilized sliding window context and HowNet lexicon to calculate attributes sentiment in concrete sentences. Kim *et al.* (2013) proposed a hierarchical aspect sentiment model (HASM) to discover a hierarchical structure of aspect-based sentiments from unlabeled online reviews to real the sentiment polarities. Liu *et al.* (2017) used heterogeneous networks and representation learning for sentiment classification of product reviews. Muhammad *et al.* (2019) developed a predictive framework utilized the semantic relations among review phrases to extract implicit and infrequent aspects for extracting and classify user's positive or negative orientation towards each aspect of online tourists' reviews.

With the development of machine learning, researchers tend to use convolutional neural network (CNN) and long short-term memory (LSTM) to quantify sentiment. In this branch of research, Cheng *et al.* (2017) used hierarchical attention networks to allocate appropriate sentiment words for given product attributes. Garcia-Pablos *et al.* (2017) built a model, namely W2VLDAm, based on word2vec (Mikolov *et al.*, 2013) and LDA for attribute category classification and sentiment polarity classification that can be adapted to any fields. Ma *et al.* (2018) proposed a model based on LSTM in which common-sense knowledge is incorporated to improve the performance of attributes extraction and sentiment polarity recognition. These models have reached a great success in attribute sentient computing.

## *2.2 Applications of Temporal Characteristics in Online Reviews*

In recent years, researchers have been regarding the temporal characteristics as an important factor in the field of online reviews mining. Hu *et al.* (2008) found that the review time of online reviews can affect product sales. The impact on sales would diminishes over time. Lee (2013) found that the most significant factor contributing to the readership was the review time of online reviews. A review posted late would lose a significant chance of being read by consumers even if it was evaluated as helpful by other users. In the work of Zhang *et al.* (2016a, 2016b), they proposed a novel approach to investigate the technology evolution by considering its social impact based on Amazon product reviews span over 18 years. Matakos and Tsaparas (2016) observed that average ratings of online reviews tend to become more polarized over time. The work did not make the purchase time of online reviews into consideration. Wu *et al.* (2016) found that coupon promotions positively impact online review ratings. However, the longer the time between the purchase and the review time, the smaller the impact. Also, Lin *et al.* (2016) found that user "purchase-review" time interval is one of the important factors affecting the quality of online comments. Shaalan and Zhang (2016) proposed a novel review weighting model on Amazon dataset combining the information on the review time and opinion quality of reviews from the online review communities to rank products. Voting time of online reviews is always useful. Wu *et al.* (2017) found that the predictive system has less prediction error when it takes the voting time information of online reviews into the model. Heydari *et al.* (2016) used the temporal characteristics to detect false comments by studying the suspicious "purchase-review" time interval series. Ning *et al.* (2015) found that the distribution of online comments in the "purchase-review" time interval was in accordance with the power-law distribution and that the relationship of power index and the user attention to the product is monotonically increasing. Li *et al.* (2015, 2016) used the Mann–Kendall trend test to divide the "purchase-review" time interval sequence into three parts, namely instant (0-2 days), medium-term (3-20 days), and long-term (21-180 days). However, this study failed to consider users' purchase time or deeply understand the evolution of user attention and sentiment toward product aspects, which limits its contribution to users. Bai *et al.* (2018) divided product lifetime into three stages, namely early, majority, and laggard stages, and then characterized and predicted reviewers purchase behavior for effective products marketing who review in the early stage. Li (2018) found that users' interest in commodities was constantly changing. The number of online comments is used to indicate the user's interest on the product. They explored whether user preferences are stable and whether interest is a dynamic based on more than 6 million Amazon online reviews. Huang et al. (2018) found that the review time presents near distance, concrete product review content is perceived as more helpful. By contrast, abstract review content is perceived as more helpful. And Camilleri (2020) found that the importance of online reviews depends on when they are presented. Murtadha *et al.* (2021) constructed a sentiment analysis model with temporal characters of online reviews that can identify changes in public sentiment over time which is important for analyzing and identifying the potential causes of a positive/negative sentiment trend at a particular point in time. However, the reserach did not make purchase time into consideration.

From the above, the temporal characteristics of online reviews are playing an important role in reviews mining. And large granularity partitioning of time intervals and incomplete time information are two of the major problems among existing studies. The paper fully considers the temporal characteristics of online reviews and utilize natural language processing technologies to explore the potential objective regularity of user attention and sentiment toward product aspects.

# 3. Methodology

The task of this paper is to explore the distribution regularities of user attention and sentiment toward product aspects. The research framework consists of three parts, as represented in Figure 2. The first part is data acquisition. The paper obtains the online reviews with temporal characteristics from JD.COM by using the reviews crawler coded by ourselves. The second part is text processing, including product aspects recognition and product aspects sentiment computing. The third part is exploring the distribution regularities of user attention and sentiment toward product aspects.

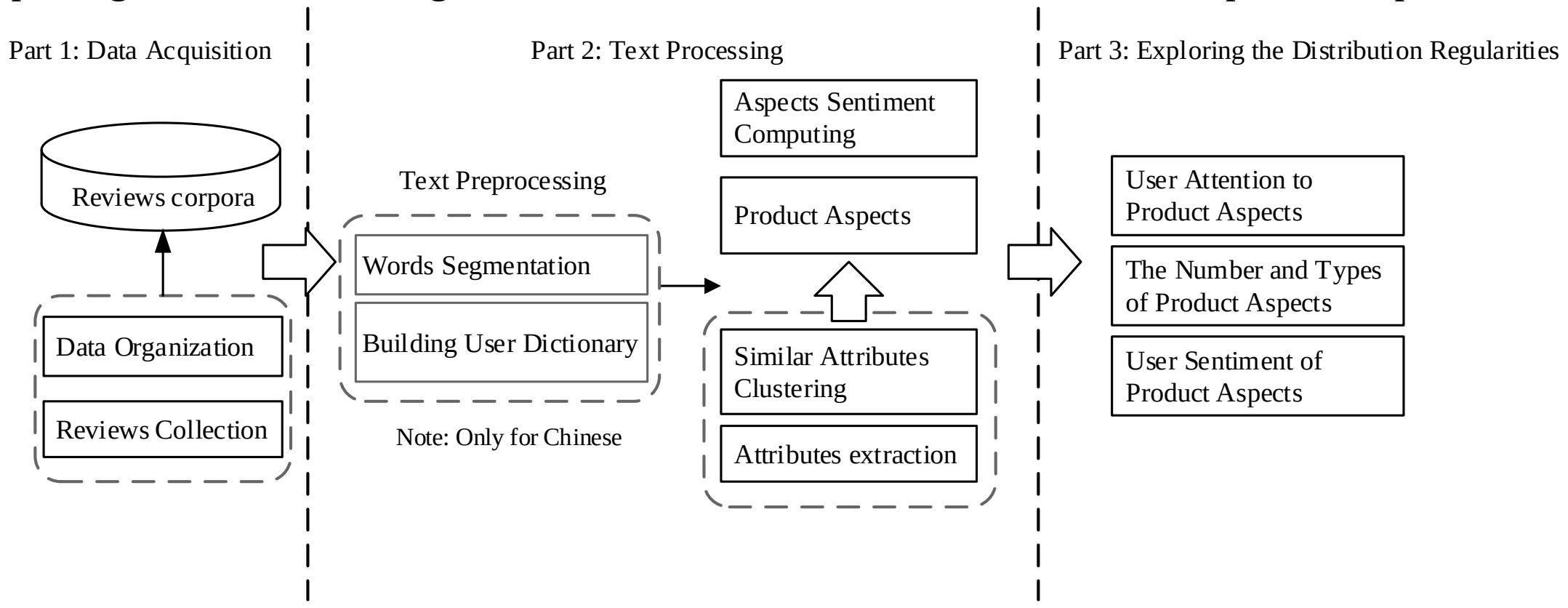

**Figure 2.** Research framework

## *3.1 Basic definitions*

Definition 1. ***User attention***: user attention to a product aspect is quantified by the count of reviews which contained the aspect. To be specific, the user attention to product aspects is described as follows:

$$Y_1 = lg(\frac{1}{k} \cdot \frac{\sum_{j\geq 1}^{k} n_{ti_C_{ij}}}{\sum_{ti=0}^{max(ti)} \sum_{j\geq 1}^{k} n_{ti_C_{ij}}}) \quad (1)$$

where $ti$ represents the time interval ($0 \le ti \le 90$ *days* in our experiments since JD.COM stipulates that users can write reviews within 90 days after purchasing), $n_{ti_C_{ij}}$ represents the number of reviews that contains the attribute $C_{ij}$ that is included in the aspect $C_i$ at the time interval $ti$, and $k$ is the number of attributes in aspect $C_i$ ($0 \le T \le 90, 1 \le j \le k$).

Definition 2. ***User sentiment***: The polarity of sentiment words corresponding to product aspect in online reviews is the user sentiment. See Section 3.4 for the specific calculation method of user sentiment.

## 3.2 Data Collection and Data preprocessing

### 3.2.1 Data Collection

The paper used the control class of WebBrowser provided by C# to develop a software to crawl online reviews from JD.COM. The raw data is organized by the format of JSON. The paper used the package of JSON to parse the raw data to extract plain text. The paper crawled more than 340k reviews contained three products, namely Samsung Galaxy S7 Edge, Huawei Mate 9, and Mi 5. The details about the datasets is shown in Table I.

**Table Ⅰ.** Details about the experimental data from JD.COM

| **Product Name** | **Count of Reviews** | **Purchase Time of Online Reviews** |
|---|---|---|
| Samsung Galaxy S7 Edge | 74,000+ | 17-Mar-2016~31-May-2017 |
| Huawei Mate 9 | 97,000+ | 28-Nov-2016~10-May-2017 |
| Mi 5 | 168,000+ | 15-Mar-2016~2-May-2017 |
| **Total** | 340,000+ | |

### 3.2.2 Data preprocessing—Building user dictionary and word segmentation

A new product typically accompanies by several new words. Since most extant dictionaries are updated slowly, and some existing rules can't be directly applied (Zhao 2012), traditional word segmentation systems of Chinese, e.g., ICTCLAS (http://ictclas.nlpir.org), are not able to identify these new words. An example is shown in Figure 3. To improve the accuracy of the segmentation system, add the new attributes that appear in the reviews to the user dictionary and annotate them with part of speech manually. And then, the paper uses ICTCLAS to segment the online reviews.

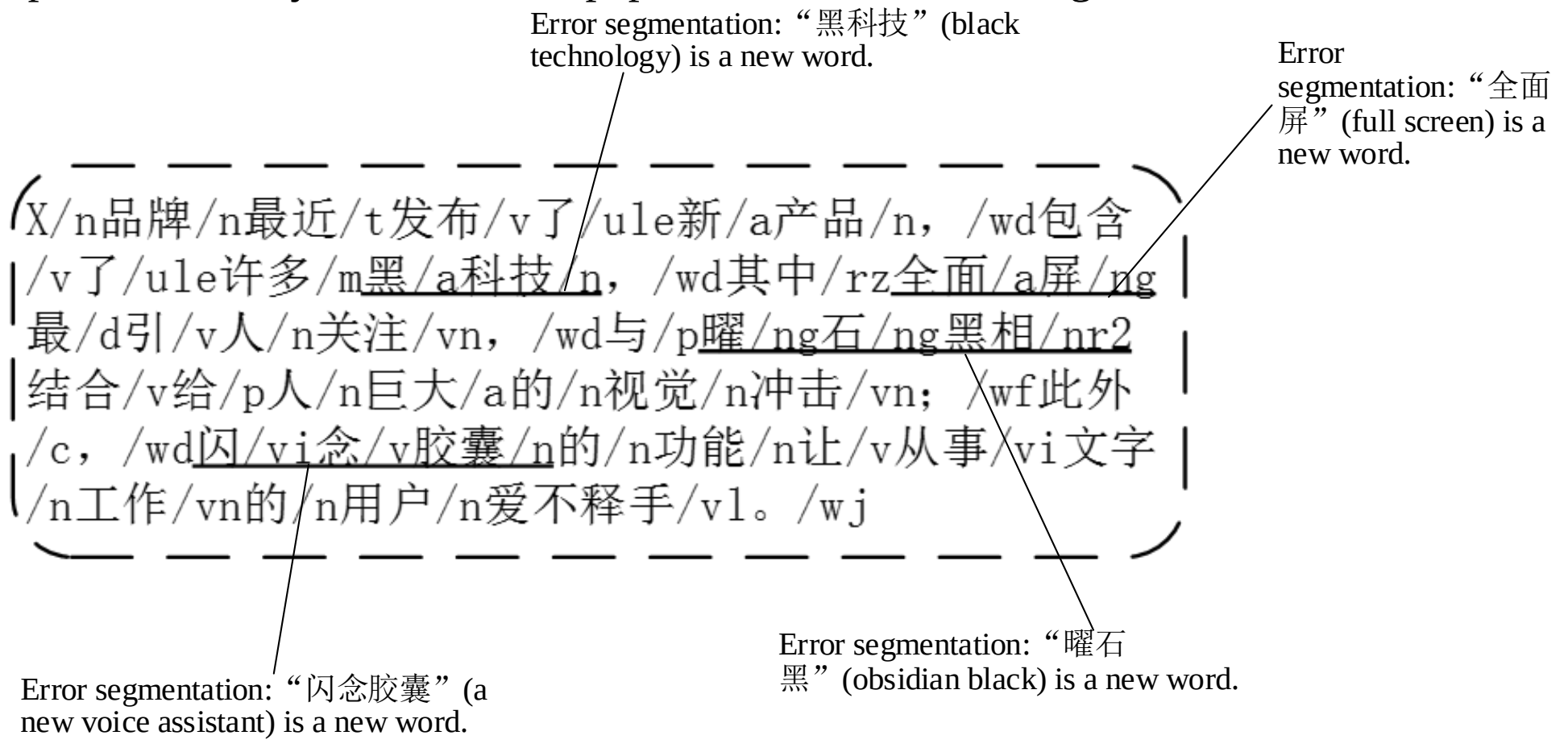

**Figure 3.** An example of segmentation results containing error segmentations by ICTCLAS

## 3.3 Product Aspects Recognition

### 3.3.1 Attributes extraction

As mentioned in 2.1.1, most traditional methods for attributes extraction are based on word frequency. It is an effective method to obtain candidate product attributes through high frequency nouns or noun phrases (Hu and Liu, 2003). For reducing the impact of incorrectly recognized product attributes, we select the noun words or noun phrases whose term frequency (TF) are over 20 as the product candidate attributes and filter the noise words by manual judgment.

### 3.3.2 Attributes clustering

There are two distinct circumstances in the current study that will produce two "similar" attributes. First, there are two or more different descriptions of one attribute in Chinese online reviews, such as "摄像头" (phone camera) and "相机" (camera), "MIUI" (an operating system based on Android) and "系统" (operating system); Second, there might be some "strong relationships," such as "电池" (battery) and "续航" (battery life), "快递师傅" (delivery man) and "物流" (logistics). The paper utilized the word2vec model (Mikolov, 2013) to generate a word vector for each attribute, denoted as $W_{aspect_i}(w_1, w_2, w_3, ..., w_n)$. The similarity of two attributes is calculated using the cosine distance (see in Formula 2). Finally, cluster the similar attributes based on the method by Qin *et al.* (2020), and then product aspects will be acquired.

$$Similarity(W_{attribute_i}, W_{attribute_j}) = \frac{\sum_{i,j=1}^{n}(W_{attribute_i} \times W_{attribute_j})}{\sqrt{\sum_{i=1}^{n} W_{attribute_i}^2} \times \sqrt{\sum_{j=1}^{n} W_{attribute_j}^2}} \quad (2)$$

The specific process of attributes clustering is divided into two stages: allocation phase and transfer phase.

***Allocation phase*:** An attribute is selected as the center of the initial cluster from the attributes set that each element has been generated word vector. In this phase, the algorithm follows a four-step procedure: (1) Traverse the elements in the set of the attributes except for the selected element as the initial center; (2) Quantify the similarity of the attributes and the center of the existing clusters; (3) Add the attribute to the cluster with the largest similarity if the similarity reaches the minimum threshold; otherwise, a new cluster is created, and the attribute is used as the center of the new cluster; and (4) Repeat (1)-(3) until each attribute of attributes set are traversed (the threshold value is determined as 0.7 in the current empirical studies).

***Transfer phase*:** Multiple clusters are generated during the allocation phase, but attributes in a cluster may have a greater value of similarity compared with a certain cluster generated later. Therefore, it is necessary to compute the similarity of the current cluster and the clusters generated later. If the elements of the current cluster have higher similarity with a certain center of clusters generated later, add the elements to the cluster and delete them from the raw cluster.

### 3.4 Aspect sentiment computing

The aim of calculating aspect sentiments is to explore the distribution regularity of user sentiment of the product aspects at different time intervals. To improve the accuracy of sentiment word matching, first of all, the NTUSD sentiment dictionary (Ku et al. 2007) should be extended because some new sentiment words are not included in the dictionary, such as the positive sentiment words "给力" (helpful, useful) and "爆表" (a high value for something) and negative sentiment words "鸡肋" (chicken ribs) and "吐槽" (complaint). The sentiment orientation value is defined as +1, while the negative sentiment word -1. The paper uses the method proposed by Ding et al. (2008) to calculate the attributes sentiment, which is based on dictionary and language rules. The attributes sentiment is calculated as:

$$S_{attribute_i} = \begin{cases} +1, \sum_{k=1}^{n} \frac{c * SO_{sw_k}}{dist(attr._i, sw_k)} > 0 \\ -1, \sum_{k=1}^{n} \frac{c * SO_{sw_k}}{dist(attr._i, sw_k)} < 0 \end{cases} \quad (3)$$

Where $n$ represents the number of sentiment words in a review, $sw_k$ represents the $k$ th sentiment word, and $SO_{sw_k}$ represents the sentiment orientation of $sw_k$. Let $c = 1$ when the first word $w_{k-1}$ and the second word $w_{k-2}$ in the front of $sw_k$ are both negative. The polarity of sentiment word $sw_k$ remains unchanged. If there exists a negative word between $w_{k-1}$ and $w_{k-2}$, let $c = -1$, and the polarity of sentiment word $sw_k$ turns over. If there is no negative word prior to $sw_k$, let c=+1. *dist* function represents the distance between an attribute and a sentiment word.

Because of the clustering of similar aspects, it is necessary to calculate the sentiment of each aspects:

$$S_{aspect} = \frac{\sum_{i=1}^{k} \sum_{j=1}^{n} S_{attribute_i}}{\sum_{i}^{k} \sum_{j=1}^{n} |S_{attribute_i}|} \tag{4}$$

Where $n$ represents the number of reviews which contain $attribute_i$, and $k$ represents the number of attributes in an aspect.

# 4. Results

Correctly extracting product attributes and accurately clustering the similar attributes are important prerequisites to explore the distribution regularities of user attention and user sentiment toward product aspects. There are 32 product aspects the paper obtain that cover almost all attributes. Taking Samsung Galaxy S7 Edge as an example, use the method mentioned in section 3.3 to obtain 156 product attributes. The extraction and clustering results are shown in Figure 4. The clustering results of Huawei Mate 9 and Mi 5 can be downloaded from https://github.com/kakabular/clustering-result.

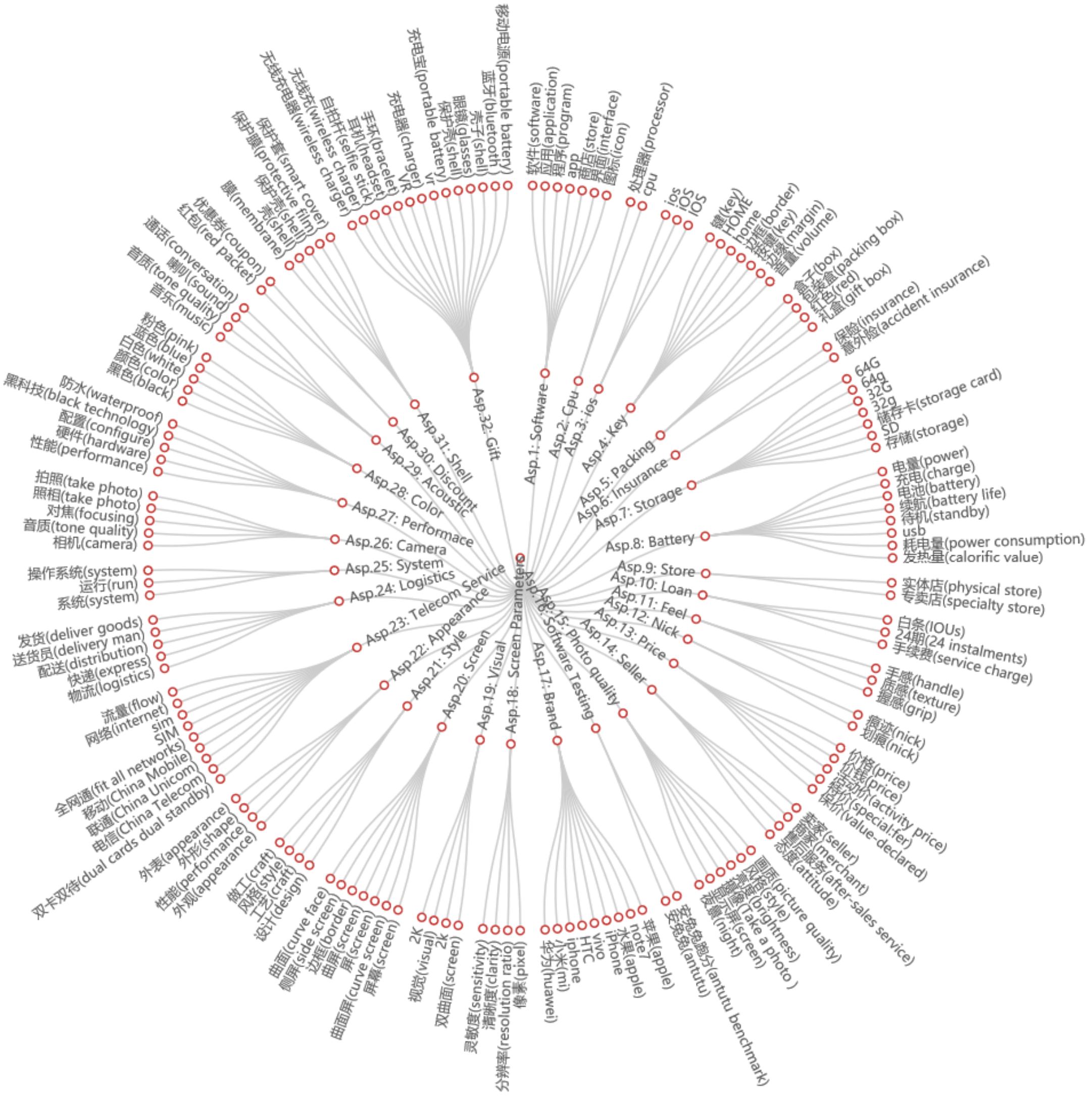

**Figure 4.** The clustering results of product attributes of Samsung Galaxy S7 Edge.

## *4.1 Exploring the distribution regularity of user attention to product aspects at different time intervals*

There will be a time interval between users' purchase time and review time. This section analyzes the distribution regularity of user attention to product aspects at different time intervals.

The user attention to product aspects at different time intervals is typically a human behavior. In recent years, through the analysis of massive data, it has been found that human behavior with temporal characteristics can be better described by the power law distribution (e.g., Zhou *et al.*, 2013). The widely used method for fitting power-law distributions is mainly based on the one-dimensional linear regression model and the least squares (Clauset *et al.*, 2012). And the paper uses $R^2$ to score the degree of fitting. The fitting processes and related formulas are represented in Appendix 1. Tables II-IV show the fitting results of aspects of three products.

**Table II** The fitting results of product aspects of Samsung Galaxy S7 Edge

| $R^2<0.6$ | | | | $0.6\le R^2<0.7$ | | | | $0.7\le R^2<0.8$ | | | | $R^2\ge 0.8$ | | | |
|---|---|---|---|---|---|---|---|---|---|---|---|---|---|---|---|
| Asp. | Int. | Coe. | **$R^2$** | Asp. | Int. | Coe. | **$R^2$** | Asp. | Int. | Coe. | **$R^2$** | Asp. | Int. | Coe. | **$R^2$** |
| | - | - | + | | - | - | + | | - | - | + | | - | - | + |
| 2 | 1.57 | 0.74 | **0.42** | *1* | 0.44 | 1.39 | **0.64** | *5* | 0.74 | 1.25 | **0.76** | *20* | 0.58 | 1.19 | **0.82** |
| 3 | 1.15 | 0.95 | **0.56** | *4* | 0.79 | 1.10 | **0.62** | *9* | 0.76 | 1.20 | **0.73** | *22* | 0.52 | 1.31 | **0.84** |
| 6 | 1.07 | 0.93 | **0.47** | *8* | 0.29 | 1.53 | **0.69** | *10* | 0.94 | 1.10 | **0.71** | | | | |
| 7 | 1.35 | 0.89 | **0.56** | *12* | 0.88 | 1.04 | **0.65** | *11* | 0.33 | 1.63 | **0.71** | | | | |
| *15* | 0.84 | 1.04 | **0.54** | *13* | 0.14 | 1.63 | **0.68** | *17* | 0.45 | 1.44 | **0.76** | | | | |
| *16* | 0.95 | 1.13 | **0.59** | *14* | 0.66 | 1.16 | **0.64** | *19* | 0.61 | 1.25 | **0.73** | | | | |
| *29* | 0.74 | 1.10 | **0.59** | *18* | 0.75 | 1.14 | **0.64** | *23* | 0.47 | 1.29 | **0.73** | | | | |
| | | | | *21* | 0.54 | 1.39 | **0.69** | *25* | 0.43 | 1.46 | **0.73** | | | | |
| | | | | *24* | 0.26 | 1.76 | **0.68** | *27* | 0.58 | 1.23 | **0.76** | | | | |
| | | | | *26* | 0.51 | 1.35 | **0.62** | *28* | 0.26 | 1.55 | **0.73** | | | | |
| | | | | *31* | 0.41 | 1.46 | **0.66** | *30* | 0.63 | 1.24 | **0.74** | | | | |
| | | | | *32* | 0.59 | 1.44 | **0.68** | | | | | | | | |

**Table III** The fitting results of product aspects of Huawei Mate 9

| $R^2<0.6$ | | | | $0.6\le R^2<0.7$ | | | | $0.7\le R^2<0.8$ | | | | $R^2\ge 0.8$ | | | |
|---|---|---|---|---|---|---|---|---|---|---|---|---|---|---|---|
| *Asp.* | Int. | Coe. | **$R^2$** | *Asp.* | Int. | Coe. | **$R^2$** | *Asp.* | Int. | Coe. | **$R^2$** | *Asp.* | Int. | Coe. | **$R^2$** |
| | - | - | + | | - | - | + | | - | - | + | | - | - | + |
| *3* | 1.14 | 1.0 | **0.53** | *1* | 0.32 | 1.44 | **0.69** | *2* | 0.61 | 1.22 | **0.70** | *28* | 0.22 | 1.57 | **0.81** |
| *6* | 1.89 | 0.56 | **0.16** | *4* | 0.87 | 1.11 | **0.69** | *7* | 0.24 | 1.49 | **0.72** | | | | |
| *10* | 0.78 | 1.18 | **0.58** | *5* | 0.28 | 1.76 | **0.60** | *8* | 0.35 | 1.39 | **0.76** | | | | |
| *16* | 1.56 | 0.8 | **0.33** | *9* | 0.86 | 1.15 | **0.67** | *11* | 0.42 | 1.47 | **0.78** | | | | |
| *18* | 0.89 | 1.11 | **0.57** | *12* | 1.02 | 1.09 | **0.66** | *17* | 0.08 | 1.65 | **0.76** | | | | |
| *30* | 1.06 | 0.98 | **0.43** | *13* | 0.71 | 1.2 | **0.69** | *19* | 0.22 | 1.53 | **0.71** | | | | |
| | | | | *14* | 0.6 | 1.2 | **0.60** | *20* | 0.71 | 1.2 | **0.71** | | | | |
| | | | | *15* | 0.85 | 1.08 | **0.60** | *21* | 0.45 | 1.45 | **0.74** | | | | |
| | | | | *22* | 0.39 | 1.51 | **0.69** | *23* | 0.51 | 1.25 | **0.70** | | | | |
| | | | | *25* | 0.11 | 1.88 | **0.63** | *24* | 0.2 | 1.76 | **0.72** | | | | |
| | | | | *27* | 0.35 | 1.45 | **0.60** | *26* | 0.37 | 1.37 | **0.75** | | | | |
| | | | | *31* | 0.88 | 1.12 | **0.64** | *29* | 0.32 | 1.44 | **0.72** | | | | |
| | | | | | | | | *32* | 0.31 | 1.50 | **0.75** | | | | |

**Table IV** The fitting results of product aspects of Mi 5

| $R^2<0.6$ | | | | $0.6\le R^2<0.7$ | | | | $0.7\le R^2<0.8$ | | | | $R^2\ge 0.8$ | | | |
|---|---|---|---|---|---|---|---|---|---|---|---|---|---|---|---|
| Asp. | Int. | Coe. | **$R^2$** | Asp. | Int. | Coe. | **$R^2$** | Asp. | Int. | Coe. | **$R^2$** | Asp. | Int. | Coe. | **$R^2$** |
| | - | - | + | | - | - | + | | - | - | + | | - | - | + |
| *3* | 1.59 | 0.76 | **0.31** | *1* | 0.53 | 1.22 | **0.66** | *5* | 0.72 | 1.23 | **0.7** | *22* | 0.49 | 1.34 | **0.80** |
| *6* | 1.64 | 0.67 | **0.20** | *2* | 0.75 | 1.14 | **0.65** | *8* | 0.48 | 1.26 | **0.78** | *28* | 0.46 | 1.36 | **0.80** |
| *19* | 1.03 | 0.99 | **0.52** | *4* | 0.27 | 1.46 | **0.67** | *9* | 0.56 | 1.25 | **0.77** | | | | |
| *30* | 0.90 | 1.04 | **0.57** | *7* | 0.46 | 1.30 | **0.68** | *10* | 1.13 | 0.94 | **0.72** | | | | |
| | | | | *12* | 0.90 | 1.08 | **0.66** | *11* | 0.45 | 1.43 | **0.71** | | | | |
| | | | | *14* | 0.61 | 1.30 | **0.67** | *13* | 0.76 | 1.07 | **0.78** | | | | |
| | | | | *16* | 0.97 | 1.12 | **0.66** | *15* | 0.84 | 1.08 | **0.73** | | | | |
| | | | | *18* | 0.77 | 1.14 | **0.68** | *17* | 0.40 | 1.33 | **0.77** | | | | |
| | | | | *21* | 1.09 | 1.00 | **0.63** | *20* | 0.69 | 1.12 | **0.71** | | | | |
| | | | | *24* | 0.28 | 1.56 | **0.69** | *23* | 0.57 | 1.19 | **0.78** | | | | |
| | | | | *25* | 0.41 | 1.37 | **0.64** | *27* | 0.51 | 1.21 | **0.72** | | | | |
| | | | | *26* | 0.49 | 1.27 | **0.65** | *32* | 0.41 | 1.34 | **0.74** | | | | |
| | | | | *29* | 0.51 | 1.24 | **0.63** | | | | | | | | |
| | | | | *31* | 0.50 | 1.40 | **0.69** | | | | | | | | |

Observed from tables above, most values of goodness of fit of user attention to product aspects are greater than 0.5. The results validate that user attention to product aspects can be fitted by pow-law distribution. Taking Samsung Galaxy S7 Edge as an example, Figure 5 shows the distribution and the power-law distribution of user attention to Aspect 20: Screen ($R^2$=0.82) at different time intervals. And the actual distribution of Aspect 20: Screen is shown in Figure 6.

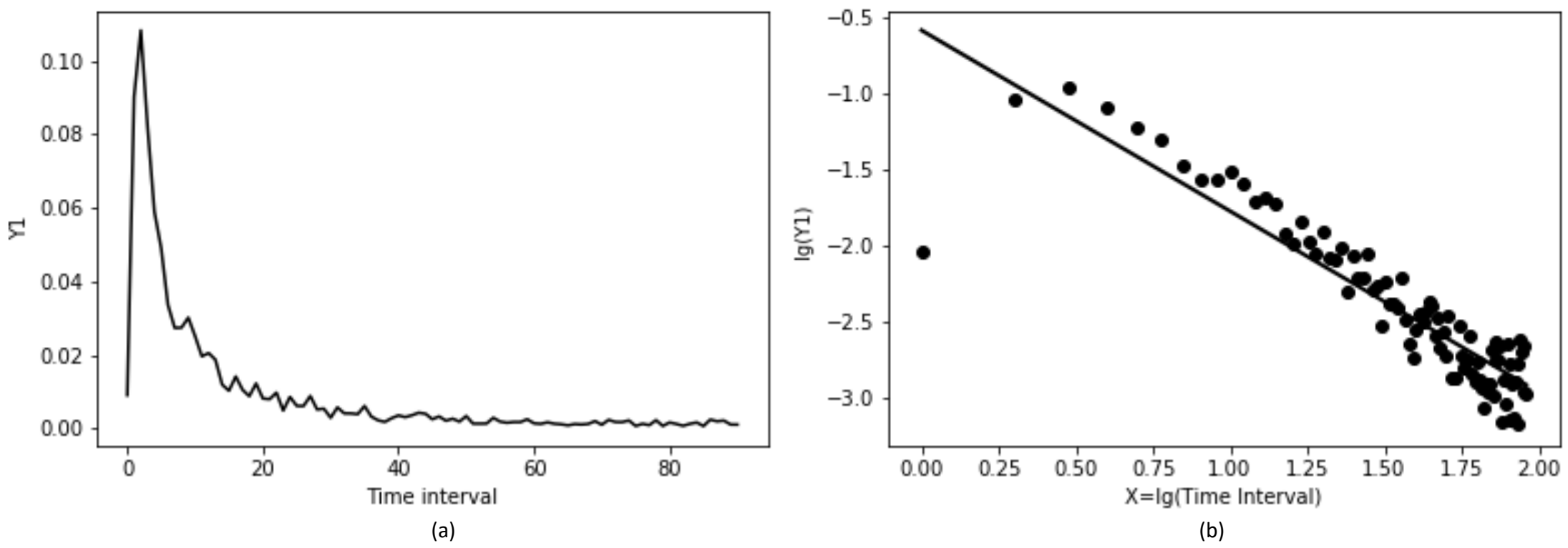

**Figure 5.** Two distributions of Aspect 20: Screen ($R^2$=0.82) from Samsung Galaxy S7 Edge: (a)The distribution of user attention at different time intervals; (b) The power-law distribution of user attention at different time intervals.

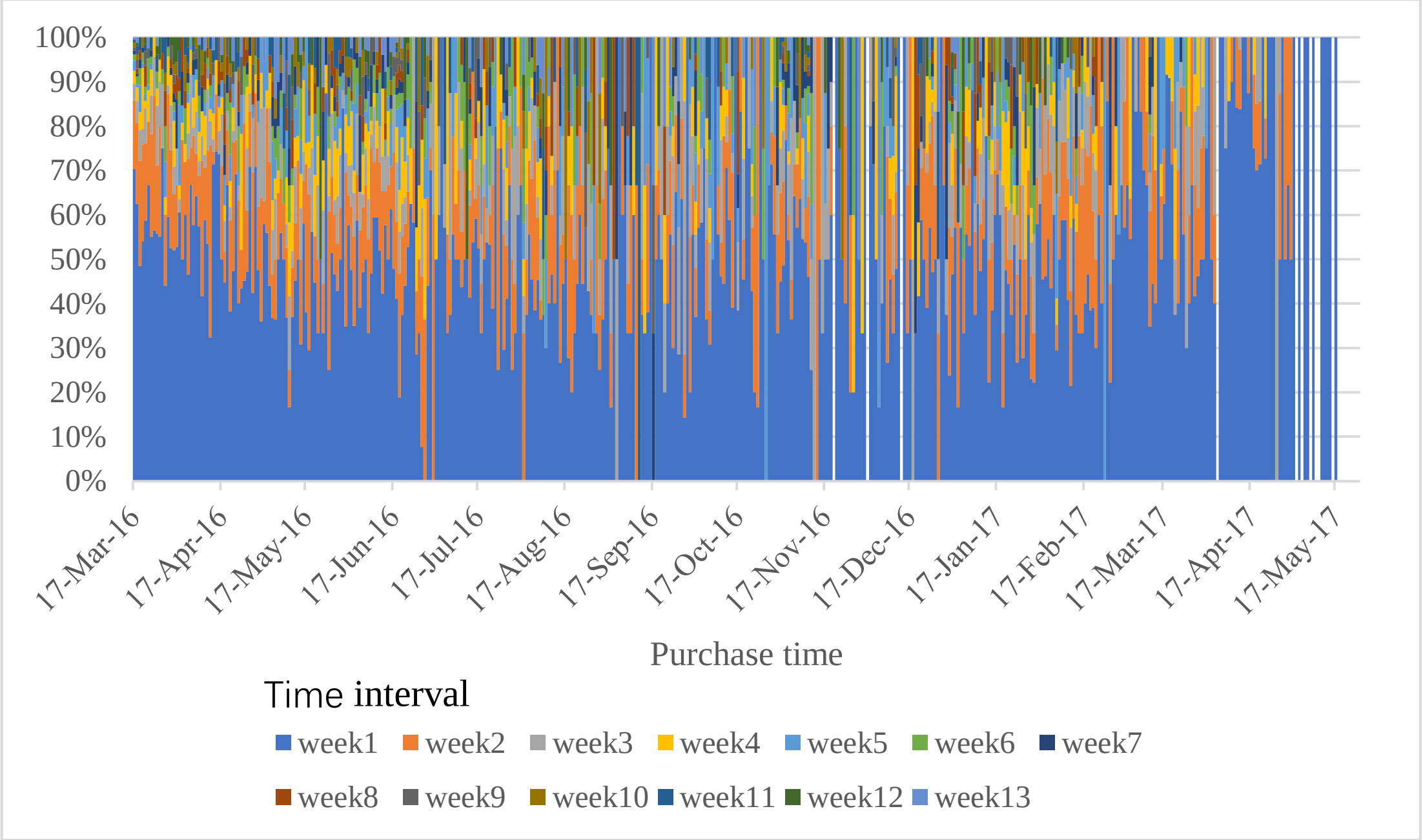

**Figure 6**. The actual distribution of Aspect 20: Screen from Samsung Galaxy S7 Edge.

## *4.2 Exploring the distribution regularity of the number and types of product aspects at different purchase time and different time intervals*

The paper uses the ratio of the number of aspects in each time interval and the count of aspects obtained (the total is 32) to represent the distribution:

$$ratio_{pt} = \frac{1}{32} \cdot n_{ti_Aspect} \tag{5}$$

where $n_{ti_Aspect}$ represents the number of product aspects in time interval $ti$. In Formula 5, 32 refers to the number of aspects clustering results. The higher the value, the more aspects it contains.

Figure 7 shows the distribution of the aspect number at different purchase time and different time intervals. From the Figure 7, it can be observed that user attention to product aspects is more comprehensive in short time intervals and have no concern with purchase time; In other words, reviews would contain more aspects in short time intervals.

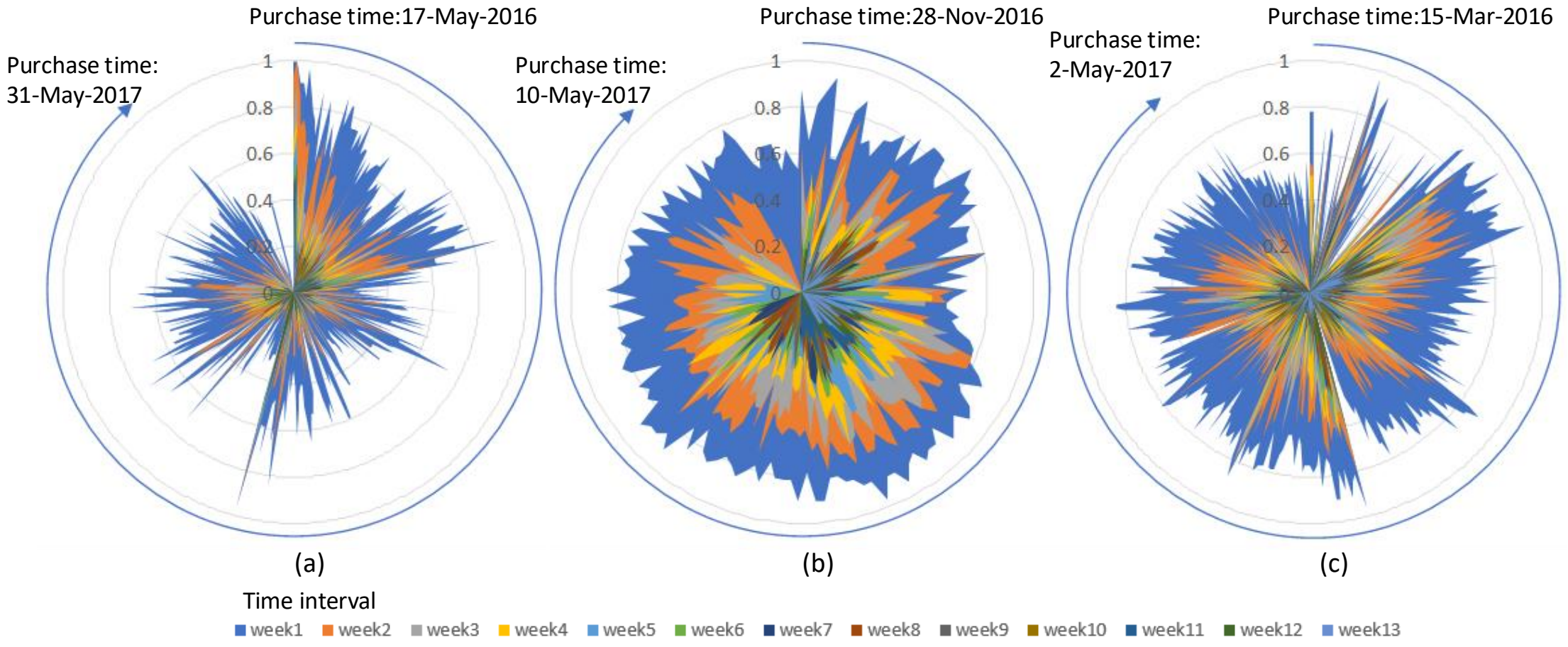

**Figure 7.** The distribution of aspects number at different purchase time and different time intervals. (a) Samsung Galaxy S7 Edge, Purchase time from 17-Mar-2016 to 31-May-2016; (b) Huawei Mate 9, purchase time from 28-Nov-2016 to 10-May-2017; and Mi 5, purchase time from 15-Mar-2016 to 2-May-2017.

The average degree of user attention to aspects is calculated as

$$Y_2 = \frac{1}{T} \cdot \sum_{ti=0}^{max(ti)} \left( \frac{1}{k} \cdot \sum_{j=1}^{k} n_{ti_C_{ij}} / \sum_{f=0}^{m} n_{ti_f} \right) \quad \textbf{(6)}$$

where $ti$ is the time interval and $0 \leq ti \leq 90$, $T = 90$, $n_{ti_C_{ij}}$ is the number of reviews that contains attribute $C_{ij}$that is in aspect $C_i$ at time interval $ti$, and $k$ is the number of attributes in $C_i$. $n_{ti_f}$ is the number of reviews that contain all attributes in different aspects at the time interval $ti$. $m$ is the number of attributes at time interval $ti$.

From Table V, users maintain a high degree of concern on the common product aspects of Screen, Appearance, System, Battery, Brand, Price, Color, Feel, Logistics and so on among three products. And it can find that users from different brand products may pay different attention to the same product aspect.

**Table V** Ten aspects of three products that users are most concerned about

| | Samsung Galaxy S7 Edge | | Huawei Mate 9 | | Mi 5 | |
|---|---|---|---|---|---|---|
| | Aspect | Average Att. | Aspect | Average Att. | Aspect | Average Att. |
| Top 1 | screen | 0.00509 | price | 0.00329 | screen | 0.00408 |
| Top 2 | appearance | 0.00455 | screen | 0.00296 | price | 0.00313 |
| Top 3 | system | 0.00391 | battery | 0.00285 | feel | 0.00312 |
| Top 4 | feel | 0.00334 | feel | 0.00267 | appearance | 0.00252 |
| Top 5 | brand | 0.00173 | logistics | 0.00245 | brand | 0.00246 |
| Top 6 | performance | 0.00169 | appearance | 0.00222 | color | 0.00207 |
| Top 7 | price | 0.00162 | performance | 0.00127 | logistics | 0.00163 |
| Top 8 | color | 0.00153 | color | 0.00119 | seller | 0.00146 |
| Top 9 | gift | 0.00145 | camera | 0.00094 | system | 0.00135 |
| Top 10 | battery | 0.00139 | gift | 0.00081 | battery | 0.00134 |

From Figure 8, it can find that almost all aspects are mentioned in short time intervals and users pay attention to several common product aspects as shown in Table 5 with the time intervals increasing. (Note that the complete illustration of Figure 8 can be downloaded from https://github.com/kakabular/clustering-result ). Select two purchase times at random, namely 17-Mar-2016 and 2-Nov-2016. The uses' attention on all product aspects at different time intervals is shown in Figure 9 (Taking Samsung Galaxy S7 Edge as an example). Taking the time intervals from week11 to week 13 as an example, it can find that uses' attention on products aspects, such as Price, Screen, Performance, Appearance, Logistics, and Brand, is significantly higher than other product aspects.

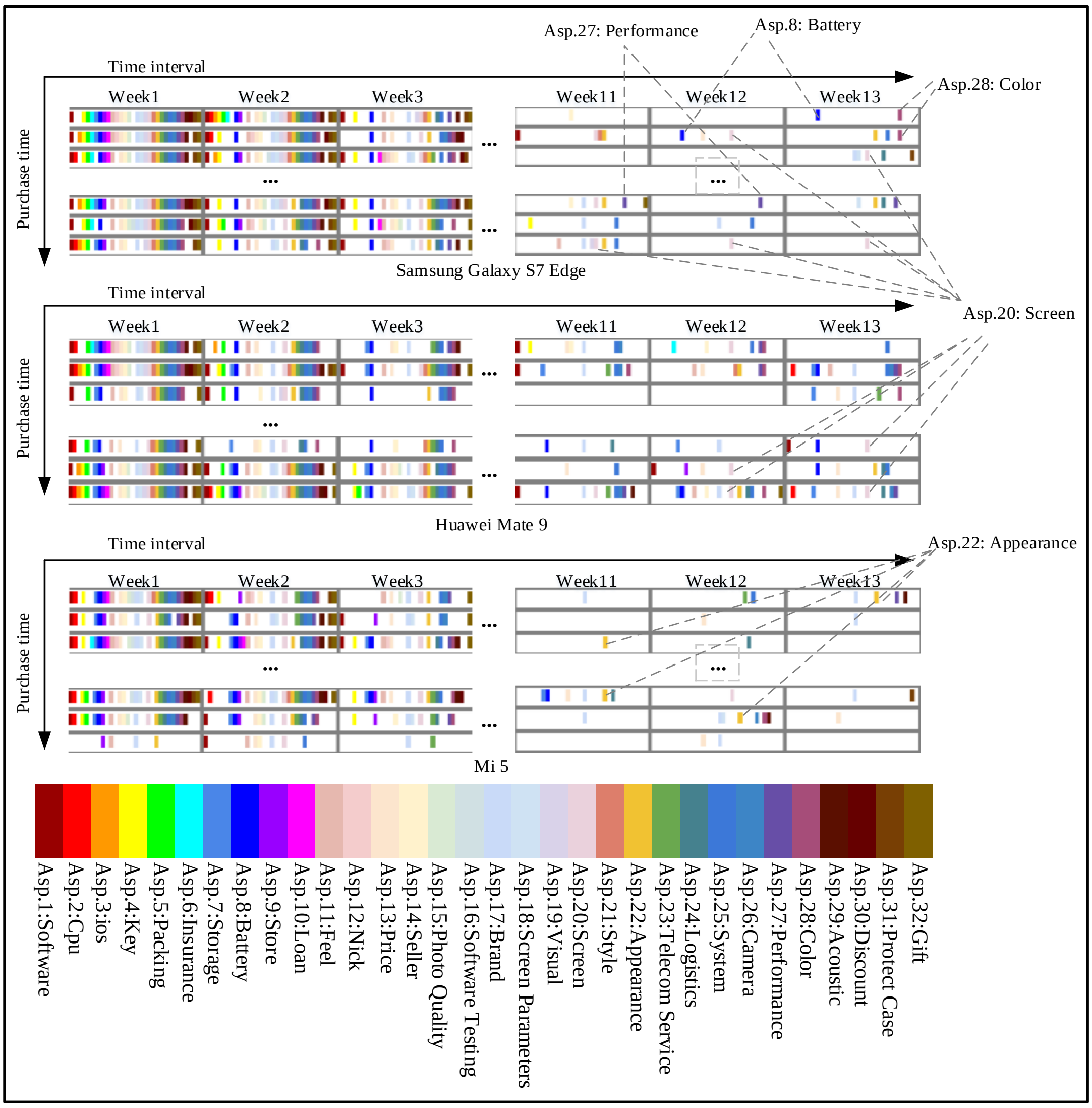

**Figure 8.** Aspect types distribution at different purchase time and different time intervals. (Note: the color order in the cells is the same as the illustration. Each cell holds 32 aspects, which are colored if they are mentioned.)

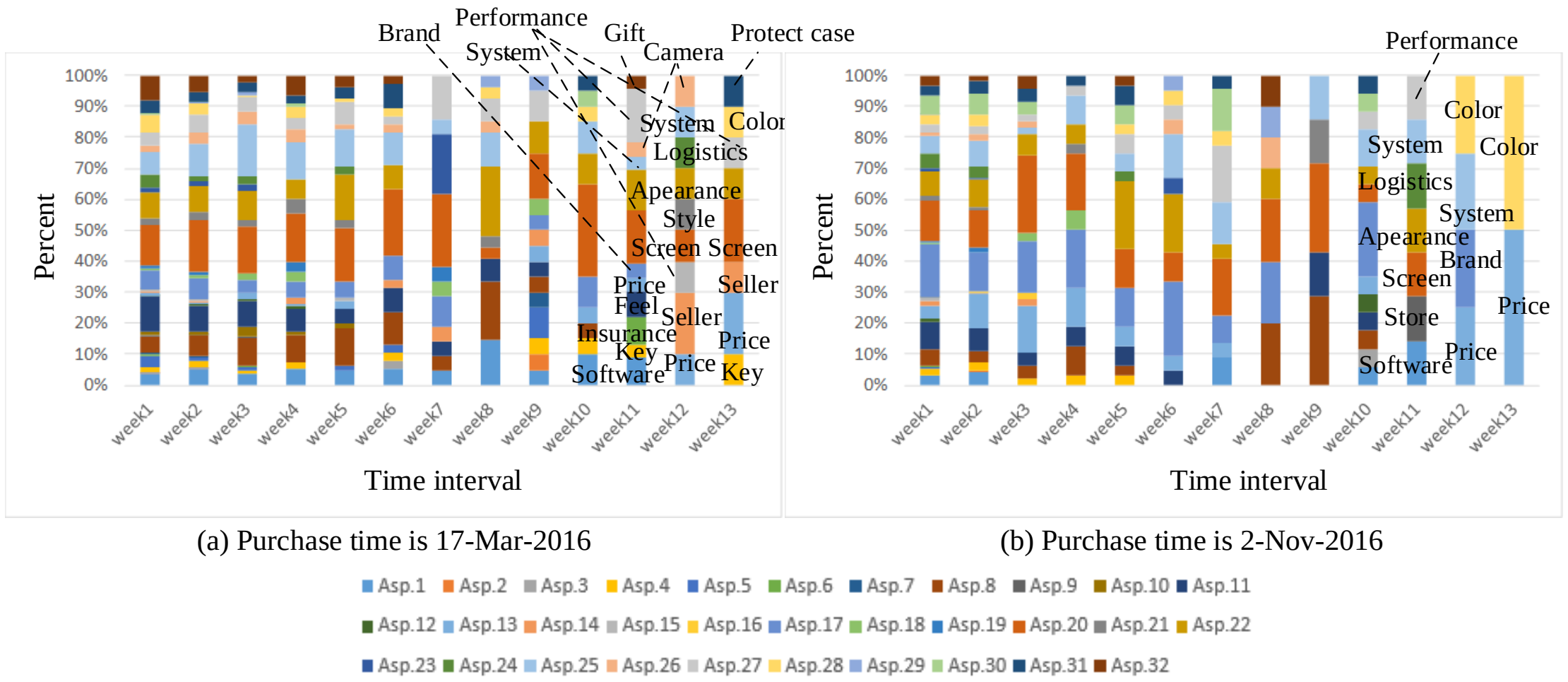

**Figure 9**. An example of user attention to all product aspects at different purchase time and different time intervals (Taking Samsung Galaxy S7 Edge as an example.)

To sum up, online reviews published in short time intervals typically contain richer information about the product aspects. In this phase, product aspects will get more user attention. With the time interval increasing, users focus on common product aspects.

### *4.3 Exploring the regularity of user sentiment of product aspects at different purchase time and different time intervals*

The distribution of user sentiment tendency of product aspects of three products at different time intervals is shown in Figure10. It can find that the tendency of user sentiment of product aspects are significantly higher/lower in short time intervals.

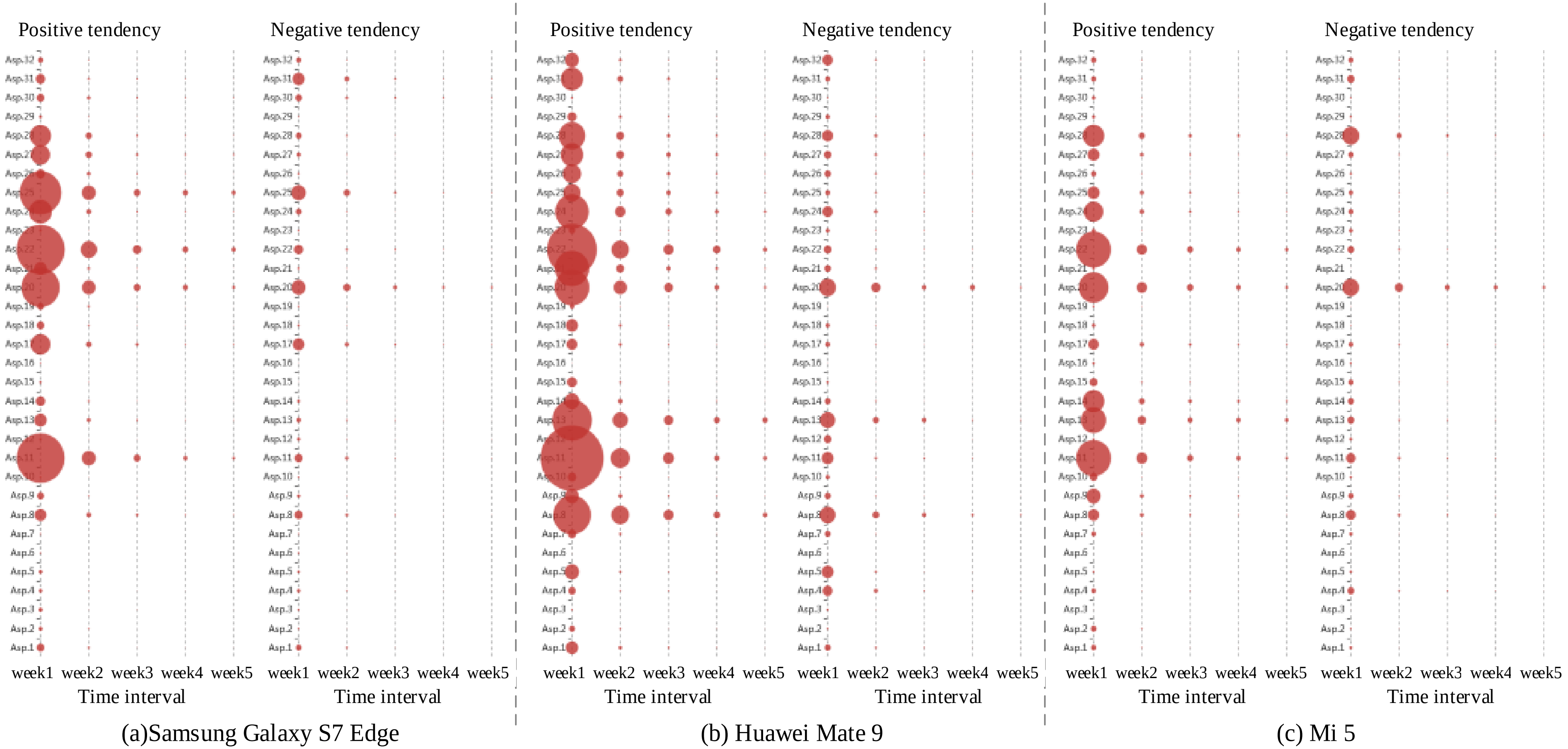

**Figure 10.** User sentiment tendency of product aspects at different time intervals

Taking Samsung Galaxy S7 Edge as an example, Figure 10 shows the distribution of user sentiment of Aspect 20: Screen at different purchase time and different time intervals. From Figure 11, it can get the same conclusion: the tendency of user sentiment of product aspects is obvious in short time intervals. Significantly sentiment can better assist users to estimate the quality of a product.

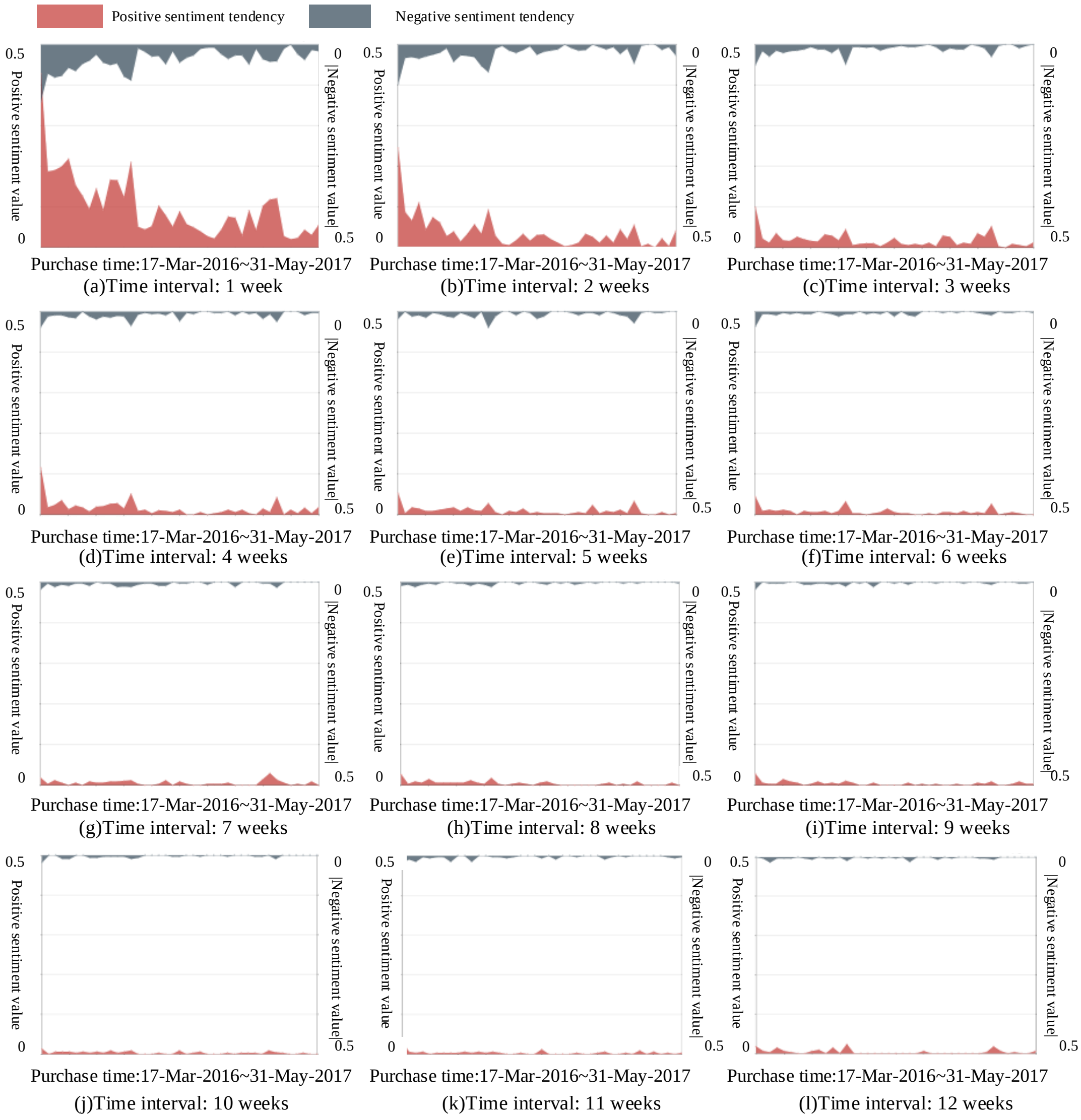

**Figure 11.** The distribution of user sentiment of Aspect 20: Screen of Samsung Galaxy S7 Edge at different purchase time and different time intervals

## 5. Discussion

This paper investigates whether there exists distribution regularities of user attention and user sentiment toward product aspects. The paper first finds that user attention to product aspects at different time intervals can be fitted by a power-law distribution. It means that product aspects will gain a lot of uses' attention in short time intervals. The finding is of great significance to product merchants. It is easy to obtain which product aspects most users pay attention to. Besides, with the time interval increases, user attention to aspects will reduce. The shopping platform should remind and encourage users to post their evaluations on product aspects to reduce the impact of missing information.

The paper also explores the distribution regularity of the number and types of product aspects at different purchase time and different time intervals. The empirical results show that the reviews published in short time intervals would contain richer product aspects and that users mostly pay attention to the common aspects as time intervals increase. Users can find more useful online reviews on product aspects in short time intervals. For shopping platforms, they need to optimize the way of online reviews display. For example, they can extract product aspects and user attention to provide a fine-grained display way for users. Besides, shopping platforms should guide users to produce comprehensive and high-quality evaluations to reduce helpless evolution and develop a method to filter the useless online reviews.

The paper also finds that users from different brand products may pay different attention to the same product aspect. For example, users from Samsung Galaxy S7 Edge pay higher attention to Screen than users from Huawei Mate 9 and Mi 5. And the priority of user attention to product aspects remain differences, which drives them to buy different brands.

Furthermore, the paper examines the distribution regularity of user sentiment of product aspects at different time intervals and find that the value of user sentiment of product aspects is significantly higher/lower in short time intervals. Potential consumers, therefore, are easy to judge the quality of a product. For merchants, the advantages and disadvantages of a product can be discovered quickly.

Generally, large-scale group behavior is regular ( Easley and Kleinberg, 2010). This paper finds that user attention to product aspects of online reviews can be fitted by a power-law distribution. Due to the lack of complete review time information in previous studies, this phenomenon has not been observed. The finding shows that the actual distribution of aspects decreases with the increase of time interval. On the other hand, online reviews contain a large number of aspects in a short time interval. Secondly, we find that even the same aspect has significant differences in user attention in different brands, indicating a preference for consumer behavior (Xue, 2020; Kim *et al.*, 2010). We also found that users' reviews posted after a short interval experience have a significant sentiment tendency. Researchers should consider online reviews ' temporal characteristics in future research related to online review aspect mining, aspect sentiment analysis, and consumer behavior analysis. This paper's findings will provide a new perspective for these studies and provide theoretical support for e-commerce platforms to optimize recommendation algorithms, optimize the display of online reviews, and improve users' shopping experience from the perspective of temporal characteristic and group preferences.

# 6. Conclusion and future work

Through the analysis of the experimental data, the paper finds that mining the product aspects of online reviews from a temporal perspective can reveal the distribution regularity of user attention and sentiment toward product aspects. Several meaningful conclusions are obtained. First, user attention to product aspects can be fitted by a power-law distribution. Second, the reviews posted in short time intervals contain more product aspects. Besides, the values of user sentiment of product aspects are significantly higher/lower in short time intervals which contribute to judging the advantages and weaknesses of a product.

However, there are several limitations in this paper. First, the experiment data is difficult to obtain because many online shopping platforms uninterruptedly upgrade themselves to constrain review crawling. Second, product aspects were mainly extracted by frequency-based, and the method cannot extract implicit aspects. Moreover, the advanced aspects extraction methods are not used in this study. Third, the sentiment dictionary used in this paper needs to be updated as time goes by. We are going to solve these deficiencies in future work.

In future work, we will focus on two main points. One is to optimize the display way of online reviews, and the other is to predict product purchase time. Potential consumers are accustomed to reading the append reviews to understand products, which means that the longer time intervals represent more profound experience and higher credibility. However, the count of append reviews is small. Can reviews at different time intervals replace append reviews? Therefore, optimizing the display way of reviews based on purchase time and time intervals can provide users with a better shopping experience. How to optimize the way of reviews display and how much better than append reviews is what the paper are going to research. Besides, purchase time of online reviews are not available in many shopping platforms, which affect the development of related researches based on purchase time. The paper plan to utilize the deep learning methods to predict the purchase time based on the data used in this paper.

# Acknowledgments

This work is supported by the National Natural Science Foundation of China (Grant No. 72074113).

## Author Bio

Chenglei Qin is currently a PhD student at Department of Information Management, School of Economics and Management, Nanjing University of Science and Technology. His research mainly focuses on text mining.

Chengzhi Zhang is a professor at Department of Information Management, Nanjing University of Science and Technology. He received his PhD degree of Information Science from Nanjing University, China. He has published more than 100 publications, including JASIST, Aslib JIM, JOI, OIR, SCIM, ACL, NAACL, etc. He serves as Editorial Board Member and Managing Guest Editor for 9 international journals (Patterns, OIR, TEL, IDD, NLE, JDIS, DIM, DI, etc.) and PC members of several international conferences (ACL, IJCAI, EMNLP, AACL, IJCNLP, NLPCC, ASIS&T, JCDL, iConference, ISSI, etc.) in fields of natural language process and scientometrics. My research fields include information retrieval, information organization, text mining and nature language processing. Currently, I am focusing on scientific text mining, knowledge entity extraction and evaluation, social media mining. He is also a visiting scholar in the School of Information Sciences (iSchool) at the University of Pittsburgh and in the Department of Linguistics and Translation at the City University of Hong Kong.

Yi Bu is an Assistant Professor at the Department of Information Management, Peking University, China. Before joining Peking University, he was a research fellow at the Center for Science of Science and Innovation (CSSI), Northwestern Institute on Complex Systems (NICO), and the Kellogg School of Management, Northwestern University. He is doing research in the application aspect of big data analytics, with a particular focus on scholarly data mining. Specifically, his research endeavors to elucidate the process of knowledge diffusion, the analysis of scholarly networks and their variants, and bibliometric indicators for research assessment. He has an undergraduate degree in information management and system from Peking University, an M.S. in data science, and a Ph.D. in informatics from Indiana University.

# Appendix: The fitting process of Power Law Distribution of user attention

The general form of power law distribution is $y = a \cdot x^{-b}$. The process is shown as follows:

(1) Taking the logarithm of both sides of the general formula simultaneously, then we obtain a formula: $\lg y = (-b) \cdot \lg x + \lg a$ . The power law distribution is a straight line whose slope is less than zero under a double logarithm coordinate;

(2) Let $Y = \lg y$ , $X = \lg x$ , $c = \lg a$ , then $\lg y = (-b) \cdot \lg x + \lg a$ is transformed into one-dimensional linear equation: $Y = (-b) \cdot X + c$ ;

(3) Put experimental data ( $x_i$ , $y_i$ ) into $X = \lg x$ and $Y = \lg y$ , respectively. Then $(X_i, Y_i)$ would be obtained, where $i = 0, 1, 2, \cdots$ ;

(4) The formula of the square of deviance is $M = \sum_{i=0}^{n} [Y_i - [(-b) \cdot X_i + c]]^2$ , then calculate the minimum extreme points $M(b, c)$ , then $b = \dfrac{(\sum_{i=0}^{n} X_i) \cdot (\sum_{i=0}^{n} Y_i) - n \cdot \sum_{i=0}^{n} (X_i Y_i)}{n \cdot (\sum_{i=0}^{n} X_i^2) - (\sum_{i=0}^{n} X_i)^2}$ , $c$ can be obtained in the same way, and then the coefficients $a$ and indices $b$ of the power law distribution could be calculated finally; and

(5) Use $R^2$ to score the degree of fitting—a value closer to one indicates that a higher goodness of fit is obtained. The formula of $R^2$ is:

$$R^2 = \frac{\sum_{i=0}^{n} [\lg(a \cdot x_i^{-b}) - \frac{1}{n} \sum_{i=0}^{n} (\lg(a \cdot x_i^{-b})]^2}{\sum_{i=0}^{n} [\lg y_i - \frac{1}{n} \sum_{i=0}^{n} (\lg(a \cdot x_i^{-b})]^2} .$$